\documentclass{article}
\usepackage{amssymb}
\usepackage[final]{corl_2025} 
\usepackage{graphicx}
\usepackage{wrapfig}
\usepackage{tikz}
\usepackage{amsmath}
\usepackage{amsthm}

\usetikzlibrary{arrows.meta,fit,calc,positioning}
\usepackage{tikz}
\usetikzlibrary{shapes.geometric, arrows.meta, positioning}
\usepackage{subcaption}
\usepackage{xcolor}
\definecolor{todoorange}{RGB}{240,120,20}

\usepackage{booktabs}
\usepackage{times} 

\usetikzlibrary{arrows.meta}

\newtheorem{definition}{Definition}
\newtheorem{proposition}{Proposition}
\usepackage{multicol,multirow}
\usepackage{algorithm}
\usepackage{algpseudocode}
\usepackage{amsmath}
\usepackage{booktabs}
\usepackage{gensymb}

\title{\textit{To Do or Not to Do}: Ensuring the Safety of Visuomotor Policies Learned from Demonstrations}
\author{
  Riad Ahmed, Moniruzzaman Akash, and Momotaz Begum \\
  Department of Computer Science \\
  University of New Hampshire \\
  \texttt{\{Riad.Ahmed, Moniruzzaman.Akash, Momotaz.Begum\}@unh.edu}
}

\begin{document}
\maketitle

\begin{abstract}
Task success has historically been the primary measure of policy performance in imitation learning (IL) research. This characteristics strictly limits the ubiquitous applications of IL algorithms in field robotics where safety assurance, in addition to task-success, is of paramount importance. It is often desirable for an IL-powered robot in the field not to roll out a policy, and hence score a poor performance, if the safety is not guaranteed. Although this trade-off between safety and performance is well investigated in classical control literature, policy safety is a heavily  underexplored domain in IL research. There is no universal definition of safety in IL. To make things worst, many existing theoretical works on safety is notoriously difficult to extend to IL-powered robots in the field. This paper offers important insights on the safety and performance of IL policies. We propose \textit{execution guarantee}, a policy-agnostic safety measure that guarantees the maximum task success for a visuomotor IL policy, despite minor run-time changes, from within a specific region in the state space. We leverage recent advances in view synthesis to identify such regions in the state space for an IL policy and explore a fundamental result on set invariance -- namely, \textit{Nagumo's sub-tangentiality condition} -- to prove and operationalize \textit{execution guarantee} from inside that region. Experiments with a Franka robot, both in simulation and real world, demonstrate how the proposed safety analysis allows various IL policies to achieve maximum task success with guarantee. We also demonstrate some interesting results on how a recovery policy -- a by-product of the proposed safety analysis -- can help to increase the policy performance and thereby mitigating the safety-performance tradeoff in IL.  
\end{abstract}
\section{Introduction}
Safety is a core requirement when it comes to bringing imitation learning (IL) algorithms (more accurately, IL-powered robots) in the field. It is however a nascent topic in IL research. The IL community has traditionally been focusing on task performance -- in terms of the number of times a policy executed the task, irrespective of the quality of the execution -- to assess learned policies. There is a large body of theoretically elegant work on safety in classical control, most of which are difficult to extend to learning-based robotic systems that operate in the real world \cite{brunke2022safe}. It is even more challenging to extend existing theoretical works on safety to IL-powered robots in the field for three primary reasons, among many others: i) IL algorithms are meant to learn a wide range of tasks from end-users; there is no general consensus on \textit{what makes an IL policy safe}. Therefore, constraints on states and/or inputs to ensure safety are not known apriori for the controller to learn during the training, ii) even if the safety is defined in a task-specific manner, safety constraints are often context dependent, making them harder to extract from demonstrations -- e.g., spilling while carrying a jar of liquid (water, milk, oil, etc.) is a serious safety threat when a robot is inside the house of an older adult but may not be equally unsafe when it is deployed in an industrial environment, iii) it is highly non-trivial to generate a reliable dynamic model of the robot and its task environment using a limited amount of noisy data that are available in field imitation learning. Note that the knowledge of the system dynamics is a prerequisite for most control-theoretic approaches on safety, such as Hamilton-Jacobi reachability \cite{bansal2017hamilton}, control barrier function \cite {ames2017cbf}, model predictive filter \cite{wabersich2022mpsf}.A handful of safety-related works that exist in the current IL literature deal with these challenges through using privileged information that are not typically available in task demonstrations -- e.g., hand designing the safety constraints for the controller to learn \cite{xiao2023safediffuser}, defining the safety criteria in a task-specific manner to avoid understanding the context \cite{generalizingSafety, fisac2018generalizing, alshiekh2018shielding}, or building a task-specific dynamic model through a simulator \cite{berkenkamp2017safe, kahn2017uncertainty} or allowing the robot to `freely interact' with a restricted task environment \cite{generalizingSafety}. Unfortunately, these strategies mar the field deployment potential of the underlying IL algorithms. The IL community has just begun the challenging journey of designing deployment-ready IL algorithms that offer theoretically-grounded yet realizable safety guarantee. This paper is among the firsts to embark on that journey. 
\par Our contributions lie in proposing a task-agnostic definition of IL policy safety and operationalizing it for visuomotor policies. We define \textit{execution guarantee} as a fundamental safety assurance that all IL policies need to offer. The \textit{execution guarantee} of a visuomotor policy is a policy-agnostic safety measure which affirms that a policy will always attain its maximum possible task success if executed from within a specific region in the state space. 
\begin{figure}
    \centering
\includegraphics[scale=0.60
]{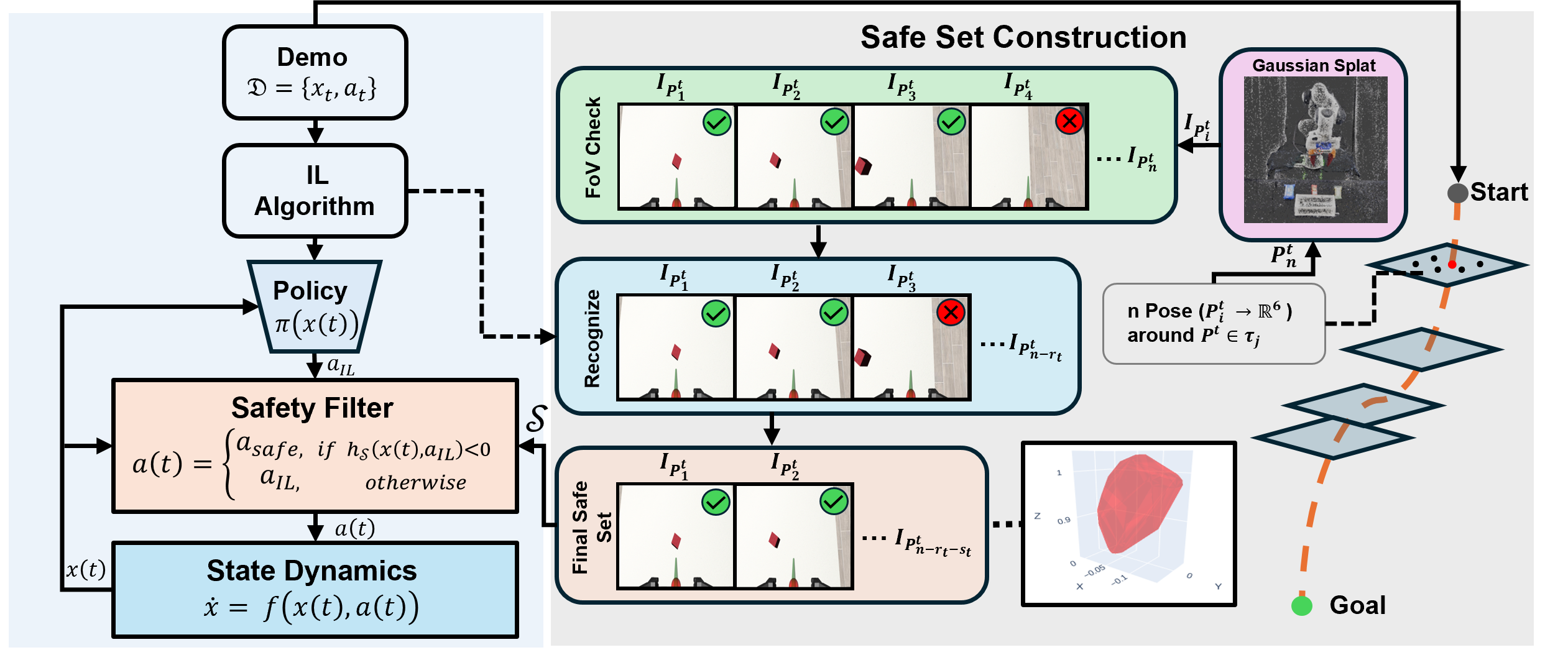}
    \caption{The proposed safety filter framework for visuomotor IL algorithms}
    \vspace*{-1.5em}
    \label{fig:overall_flow}
\end{figure}
We leverage recent advances in computer vision research on novel view synthesis to identify such regions in the state space for IL policies and explore a fundamental
result on set invariance -- namely, \textit{Nagumo's tangentiality condition} published in 1942 \cite{nagumo1942} -- to prove that the region is indeed a control invariant (CI) set for a system whose trajectory is evolving under the IL policy. This CI set synthesis process only leverages the demonstration data that the policy was trained on and do not make any assumption about the dynamic/world model of the robot. Finally, devise a recovery controller that can always keep a learned IL policy within the CI set from where the \textit{execution guarantee} exists. Experiments with a Franka robot, both in simulation and real world, show the effectiveness of the proposed work in allowing various IL policies to achieve maximum task success with guarantee. Note that the proposed work is only applicable for IL from \underline{visual demonstrations}. If an IL policy has a dedicated vision processing pipeline, the construction of the CI set takes that into consideration to maintain \textit{execution guarantee}.  

\section{Related Work}


\paragraph{Safety in Imitation Learning}While imitation learning (IL) has enabled impressive robotic capabilities, its integration with safety remains limited. Most IL approaches, such as behavior cloning, lack robustness to distributional shifts and do not account for failure cases at deployment. A few recent works have attempted to incorporate safety explicitly. SafeDagger~\cite{safedagger} improves robustness by incorporating expert interventions during unsafe rollouts, but it requires online supervision and is impractical for autonomous field use. SafeDiffuser~\cite{xiao2023safediffuser} integrates barrier conditions into diffusion-based policies, but assumes simulator access and relies on handcrafted safety conditions. RAIL~\cite{jung2024rail} constrains policies using latent reachability sets but primarily captures geometric feasibility, with no integration of perceptual constraints. These works mark progress, but they remain reliant on privileged information or retraining, which limits scalability and generalization.
\vspace*{-1.0em}
\paragraph{Safety Beyond Imitation Learning}In classical control and reinforcement learning, safety has been more extensively studied. Control barrier functions (CBFs)~\cite{ames2017cbf, blanchini1999set} and Hamilton–Jacobi reachability analysis~\cite{bansal2017hamilton} offer formal guarantees by ensuring forward invariance or defining safe reachable sets. These methods are theoretically sound but depend on known system dynamics and structured environments. More recent approaches extend safety frameworks into the visual domain—e.g., Tayal et al.~\cite{tayal2024semi} learn barrier functions from raw images, while ABNet~\cite{xiao2024abnet} and BarrierNet~\cite{xiao2023barriernet} generate vision-based safety certificates via neural attention and transformers. Latent-space methods~\cite{generalizingSafety, wabersich2023data} define reachable or invariant sets using learned embeddings but often rely on simulator-generated data or policy retraining to function reliably.
\vspace*{-1.0em}
\paragraph{Applicability to IL Settings}Despite the richness of safety literature, most existing methods are not directly applicable to field-deployable IL systems. Classical safety approaches such as control barrier functions~\cite{ames2017cbf, blanchini1999set} and reachability-based methods~\cite{bansal2017hamilton} assume known system dynamics and structured environments, which are rarely available in the field. Vision-based safety methods~\cite{tayal2024semi, xiao2024abnet} often require full control over the training pipeline or involve retraining policies to learn barrier certificates. In contrast, IL is typically used in low-data, high-variability scenarios, where policies are trained once from demonstrations and deployed in unstructured, partially observed environments~\cite{osa2018algorithmic}. Our work fills this gap by introducing a deployment-ready safety framework that operates entirely offline, requires no access to the environment model or policy internals, and leverages the policy’s own perception system to define a semantically meaningful safeset. This enables runtime enforcement of safety without retraining, making it well-suited for real-world IL applications~\cite{brunke2022safe, xiao2023safediffuser}.
\vspace*{-0.5em}
\section{Preliminaries}
\subsection{IL-Powered Robot as a Dynamic System} 
A robotic system (e.g., a manipulator) executing a learned policy \( \pi \) can be modeled as a continuous-time dynamical system: \( \dot{x}(t) = f(x(t), a(t)) \), where \( x(t) \in \mathbb{R}^{n_x} \) denotes the state and \( a(t) \in \mathbb{R}^{n_a} \) the control input at time \( t \in \mathbb{R}_{\geq 0} \), with \( \pi : x(t) \rightarrow a(t) \).

For many IL-based manipulators, a simplified fully-actuated model is assumed: \( \dot{x}(t) = g(x(t)) a(t) + A \), where \( g(x(t)) \in \mathbb{R}^{n_x \times n_a} \) maps control to state-space velocities and \( A \) is a constant offset vector. This abstraction allows us to reason about set invariance and safety constraints in pose space directly.


\vspace{-0.3em}
\noindent\textbf{Assumption 1.} \textit{No obstacle occludes the goal object or configuration at runtime unless it was present in the demonstrations.}

\vspace{-0.3em}
\noindent\textbf{Proposition 1.} \textit{A fully actuated robot governed by an IL policy can execute any obstacle-free, line-of-sight trajectory connecting the current and goal states.}

\vspace{-0.3em}
\noindent\textit{Remarks.} Assumption 1 is standard in IL settings. Proposition 1 holds under the condition that both current and goal states lie near the distribution of demonstration data. Execution efficiency (e.g., path optimality or smoothness) depends on the IL algorithm; for instance, diffusion-based policies \cite{diffusion} typically outperform basic behavior cloning \cite{mandlekar2021matters}.

\vspace*{-0.5em}
\subsection{Set Invariance and Tangent Cones}
\label{sec:set_invariance}

Set invariance is foundational to safety filter design \cite{blanchini2008set, brunke2022safe}. We briefly review key definitions:

\begin{definition}[Positive Invariance]
A set \( \mathcal{S} \subseteq \mathbb{R}^{n_x} \) is positively invariant if \( x(0) \in \mathcal{S} \) implies \( x(t) \in \mathcal{S} \) for all \( t \geq 0 \).
\end{definition}

\begin{definition}[Control Invariance]
A set \( \mathcal{S} \subseteq \mathbb{R}^{n_x} \) is control-invariant (weakly positively invariant) if there exists at least one control input keeping \( x(t) \in \mathcal{S} \) for all \( t \geq 0 \).
\end{definition}

\begin{definition}[Tangent Cone]
For a closed set \( \mathcal{S} \), the tangent cone at \( x \in \mathcal{S} \), denoted \( \mathcal{T}_\mathcal{S}(x) \), is the closure of all directions in which the system can move while remaining in \( \mathcal{S} \).
\end{definition}
\vspace*{-0.8em}
\noindent\textit{Remark.} The tangent cone is nontrivial only at the boundary \( \partial \mathcal{S} \); inside \( \mathcal{S} \), it equals \( \mathbb{R}^{n_x} \).
\vspace*{-0.8em}
\subsection{Nagumo’s Theorem}
\label{sec:nagumo}

Nagumo's Sub-tangentiality condition provides a necessary and sufficient criterion for set invariance in continuous-time systems:

\begin{definition}[Nagumo’s Sub-tangentiality Condition \cite{nagumo1942}]
A set \( \mathcal{S} \) is control-invariant under the dynamics \( \dot{x}(t) = f(x(t), a(t)) \) if and only if:
\(
\dot{x}(t) \in \mathcal{T}_\mathcal{S}(x(t)), \quad \forall x(t) \in \partial \mathcal{S}.
\)
\end{definition}
\vspace*{-0.8em}
\noindent\textit{Remark.} Geometrically, this condition requires that at the boundary of \( \mathcal{S} \), the system's velocity vector points into or tangent to the set.

\vspace*{-0.8em}

\section{Problem Definition}
\label{sec:problem}

The goal of this work is to equip imitation learning (IL) policies with a principled form of safety assurance that enables reliable deployment in real-world settings. A key challenge in doing so is defining what it means for an IL policy to be ``safe,'' especially in the absence of explicit dynamics models or known constraints. We address this by introducing the notion of an \textit{execution guarantee}—a task-agnostic, policy-aware definition of safety for IL.

\begin{definition}[Execution Guarantee]
Given a set of demonstrations \( \mathcal{D} \) and an IL policy \( \pi \) trained on \( \mathcal{D} \), we say that \( \pi \) satisfies an \textit{execution guarantee} from a set \( \mathcal{S} \subseteq \mathbb{R}^{n_x} \) if the system, when initialized in \( \mathcal{S} \) and controlled by \( \pi \), achieves its maximum attainable task success rate.
\end{definition}

\noindent\textbf{Remark.} Execution guarantee does not imply that \( \pi \) is optimal; rather, it ensures that the policy achieves its best-known performance—typically observed during evaluation on demonstration-aligned conditions—when the system operates within a specific region \( \mathcal{S} \). Crucially, this notion abstracts away from model-specific details: the guarantee does not require access to the system dynamics or environment model. A trivial lower bound on \( \mathcal{S} \) is the set of states explicitly visited in \( \mathcal{D} \), but this is typically a strict subset of the full region from which the policy can succeed. An accurate, possibly expanded estimate of \( \mathcal{S} \) is therefore essential for certifying safe policy execution in practice.To operationalize this definition, we define task success via a goal set \( \mathcal{X}_{\text{goal}} \subseteq \mathbb{R}^{n_x} \), representing states where the task is considered complete. For example, in a pick-and-place task, \( \mathcal{X}_{\text{goal}} \) may include states where the end-effector is grasping the correct object. This set can be extracted from demonstrations using existing goal inference methods (e.g., \cite{sojib2024self}). We formalize the execution guarantee as follows:
\begin{equation}
    x(t) \in \mathcal{S} \; \forall t \geq 0 \quad \Rightarrow \quad \exists t' > 0: \; x^{\pi}(t') \in \mathcal{X}_{\text{goal}}.
    \label{eq:execution_guarantee}
\end{equation}
\vspace{-0.2em}
That is, if the system remains within \( \mathcal{S} \) under policy \( \pi \), it is guaranteed to eventually reach a goal state. Our objective is twofold: (1) to identify conditions under which a given set \( \mathcal{S} \) satisfies Eq.~\ref{eq:execution_guarantee}, and (2) to construct such a set for a pretrained visuomotor policy using only demonstration data and the policy's perception module. The next section introduces our framework for addressing both goals.

\section{A Set-Theoretic Framework for Evaluating Execution Guarantee}
\label{sec:framework}

We now present a set-theoretic framework for evaluating and enforcing \textit{execution guarantee} for imitation learning (IL) policies. The approach builds on the concept of control-invariant sets and leverages Nagumo's sub-tangentiality condition to certify safe regions from which the policy reliably succeeds. We begin with the practical assumption that the IL policy \( \pi \) incorporates or is paired with a known perception module. Many modern visuomotor policies include built-in perception (e.g., diffusion-based \cite{diffusion}, transformer-based \cite{arnab2021vivit}), while others use modular visual front-ends. Our central observation is as follows:

\begin{proposition}
There exists a subset of the state space—beyond the demonstration data—where the task goal remains both (i) visible and (ii) recognizable by the policy's perception system.
\end{proposition}

\noindent\textit{Remark.} Under Assumption 1, this proposition holds due to the continuity of sensory data in natural scenes. When the policy remains in perceptually familiar regions, it remains effectively in-distribution, increasing the likelihood of correct execution. However, identifying such regions is nontrivial and policy-dependent.

Let \( \mathcal{S} \subset \mathbb{R}^{n_x} \) denote the set of states that satisfy the conditions in Proposition 1 for a given policy \( \pi \). We define a scalar function \( h : \mathbb{R}^{n_x} \rightarrow \mathbb{R} \) that characterizes this set via its superlevel set:
\begin{equation}
    h(x) > 0 \;\Leftrightarrow\; x \in \text{int}(\mathcal{S}), \quad h(x) = 0 \;\Leftrightarrow\; x \in \partial \mathcal{S},
    \label{eq:h_levelset}
\end{equation}
where \( h(x) \) is Lipschitz continuous and differentiable, with \( \nabla h(x) \neq 0 \) on the boundary \( \partial \mathcal{S} \), and \( \text{int}(\mathcal{S}) \neq \emptyset \).

Given this structure, and assuming the robot is fully actuated (see Proposition 1), there always exists a control input \( a' \) such that the induced velocity \( \dot{x}(t) \) satisfies Nagumo’s sub-tangentiality condition:
\begin{equation}
    \nabla h(x)^\top \dot{x}(t) \geq 0, \quad \forall x \in \partial \mathcal{S}.
    \label{eq:nagumo_satisfied}
\end{equation}
This condition certifies that \( \mathcal{S} \) is a control-invariant set for the dynamics in \( \dot{x(t)} \), and that the policy can be kept within \( \mathcal{S} \) indefinitely using an auxiliary \textit{recovery control signal} \( a' \). Since \( \mathcal{S} \) is composed of perceptually valid states (by construction), trajectories under \( \pi \) initiated in \( \mathcal{S} \) are guaranteed to reach the goal set \( \mathcal{X}_{\text{goal}} \), thereby satisfying Eq.~\ref{eq:execution_guarantee} and ensuring an execution guarantee.

\subsection{Operationalizing the Safety Function \( h(x) \)}
\label{sec:operationalize_hx}

To construct \( \mathcal{S} \), we define \( h(x) \) such that it is positive only when both perceptual criteria are met:
\begin{equation}
    h(x) \geq 0 \;\; \text{iff} \;\; \mathcal{X}_{\text{goal}} \in \text{FOV}(x) \; \text{and} \; \mathcal{X}_{\text{goal}} \in \text{RECOG}(x),
    \label{eq:h_visibility}
\end{equation}
where \( \text{FOV}(x) \) evaluates whether the goal is visible from state \( x \), and \( \text{RECOG}(x) \) measures whether the policy can interpret visual observations from that pose. States failing either condition are assigned negative \( h(x) \), forming the exterior of the safeset.

\subsection{Constructing \( h(x) \) for Visuomotor Policies}
\label{sec:h_x}

Visuomotor policies typically operate in a 7D state space (\( n_x = n_a = 7 \)) comprising 3D position and 4D orientation (quaternion). We construct \( h(x) \) in this space via discretization and linear interpolation, using the following three-stage pipeline:
\vspace*{-1.0em}
\paragraph{Step 1: Candidate State Generation.} 
We perturb demonstration trajectories in both translation and rotation to explore the vicinity of expert trajectories. Translational noise is sampled in spherical coordinates relative to the goal, while rotational perturbations are added via small Euler-angle offsets. Sampling is denser near the goal, where perception sensitivity is highest. Each perturbed state is a 7D vector in SE(3).
\vspace*{-1.0em}
\paragraph{Step 2: Visibility Filtering via \( \text{FOV}(x) \).}
We compute a continuous visibility score that decays as the goal shifts toward the image periphery, becomes occluded, or exits the camera frustum. This is done using camera intrinsics and known geometry to evaluate projective visibility. The resulting FOV score is Lipschitz continuous with respect to \( x \), and enables gradient-based boundary characterization.
\vspace*{-1.0em}
\paragraph{Step 3: Recognizability Filtering via \( \text{RECOG}(x) \).}
We freeze the policy’s visual encoder and treat it as a feature extractor. Embeddings from demonstration-time views define a reference distribution in feature space. At each candidate state, we render an image using a view synthesis model, compute its embedding, and evaluate a Mahalanobis distance to the reference set. This is converted into a soft recognizability score via exponential decay, also ensuring Lipschitz continuity.

\begin{figure}[t]
\centering
\begin{minipage}[t]{0.52\textwidth}
\centering
\includegraphics[width=\linewidth]{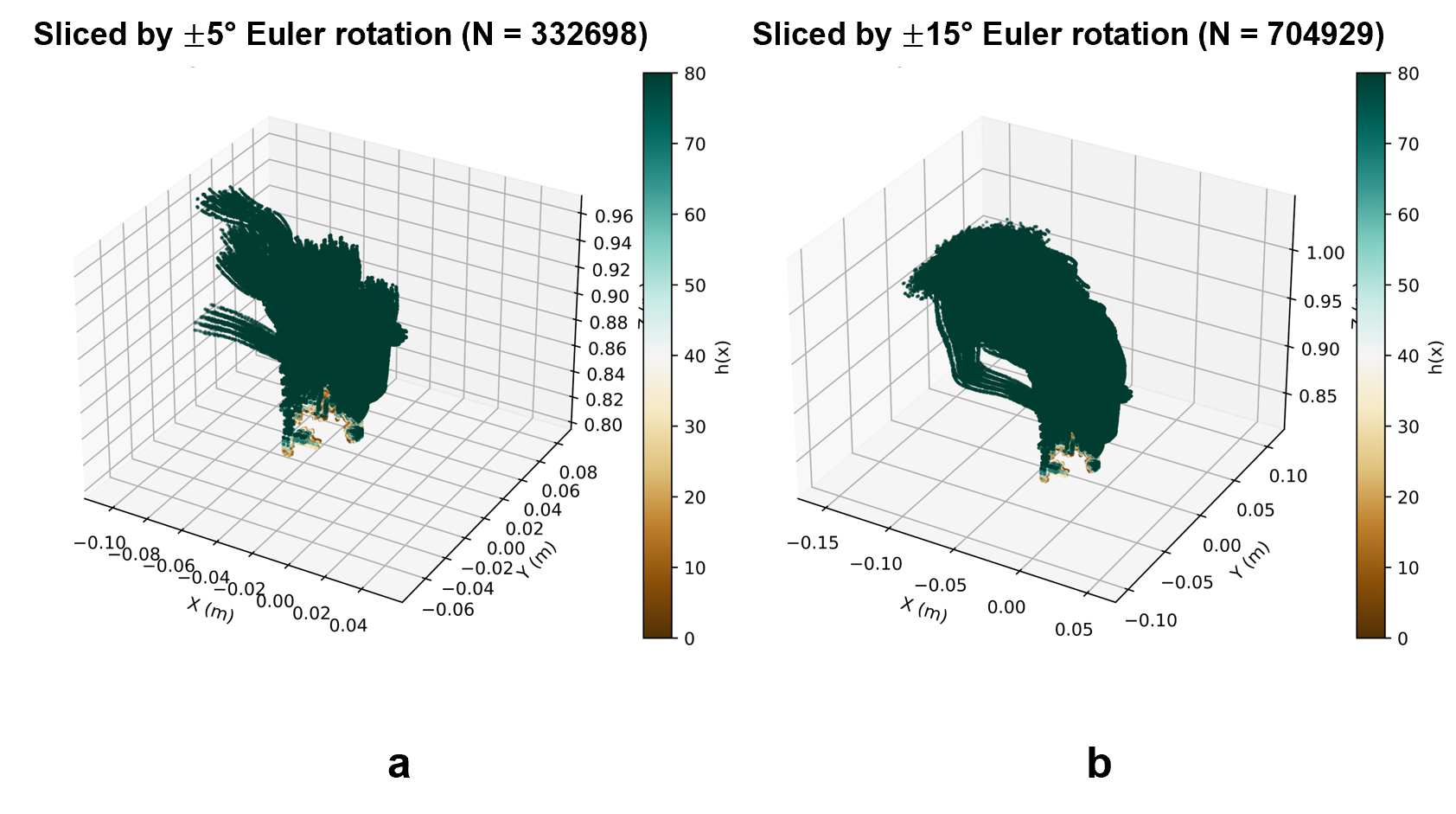}
\vspace*{-1.5em}
\captionof{figure}{Construction of the control-invariant set $\mathcal{S}$. (a) Zero super-level set of $h(x)$ at $\pm5^\circ$. (b) Same set at $\pm15^\circ$}
\label{fig:safesets}
\end{minipage}
\hfill
\begin{minipage}[t]{0.45\textwidth}
\vspace*{-11.2em} 
\scriptsize
\captionof{algorithm}{Construction of Safe Set and Recovery Controller}
\label{alg:safeset_recovery}
\begin{algorithmic}[1]
\Require Policy $\pi(x)$; demonstrations $\mathcal{D} = \{\tau_i\}_{i=1}^N$
\Ensure Safe set $\mathcal{S}$; recovery controller $a'(x)$
\State \textbf{Candidate Generation:}
\For{each $x$ in $\mathcal{D}$}
    \State Sample states $\{x^{(j)}\} \gets \texttt{Sample}(x)$
\EndFor
\State $\mathcal{C} \gets \bigcup \{x^{(j)}\}$
\State \textbf{Evaluate $h(x)$:}
\For{$x \in \mathcal{C}$}
    \State $h(x) \gets \text{fov}(x) \cdot \text{recog}(x)$
\EndFor
\State $\mathcal{S} \gets \{x \in \mathcal{C} \mid h(x) \geq 0\}$
\State $a'(x) = \arg\min_u \|u - \pi(x)\|^2 \text{ s.t. } \nabla h(x)^\top u \leq \alpha h(x), \alpha > 0$
\State \Return $\mathcal{S},\ a'(x)$
\end{algorithmic}
\end{minipage}
\end{figure}

Figure~\ref{fig:safesets} visualizes the safeset construction pipeline: positive \( h(x) \) values indicate perceptually confident regions, and the zero-level set forms \( \mathcal{S} \), the safe execution set.

\subsection{Recovery Control via Safety-Constrained Projection}
\label{sec:recovery}

Once \( h(x) \) is defined and \( \mathcal{S} = \{x \mid h(x) \geq 0\} \) is constructed, we enforce trajectory containment via a recovery control signal \( a' \). Using the Nagumo condition, we define \( a' \) as the solution to the following Quadratic Program (QP):
\begin{equation}
    a' = \arg\min_{u} \| u - u_{\pi} \|^2 \quad \text{s.t.} \quad \nabla h(x)^\top u \leq \alpha h(x),
    \label{eq:recovery_qp}
\end{equation}
where \( u_{\pi} \) is the nominal policy output and \( \alpha > 0 \) is a tunable margin encouraging early recovery before reaching \( \partial \mathcal{S} \). This QP minimally perturbs the learned action while maintaining set invariance, thus enforcing the execution guarantee in practice.
\newcommand{\COMMENT}[1]{\hfill$\triangleright$ #1}

\section{Experimental Evaluation}
\label{sec:experiments}
\vspace*{-1em}
We evaluate our proposed safety framework through comprehensive experiments in both simulation and real-world environments. The evaluation focuses on three core objectives: (1) validating that learned control-invariant (CI) sets \( \mathcal{S} \) enable safe and successful policy execution; (2) assessing the effectiveness of the recovery controller \( a' \) in maintaining trajectories within \( \mathcal{S} \); and (3) demonstrating that policies constrained to \( \mathcal{S} \) are robust to significant out-of-distribution (OOD) disturbances. We adopt two state-of-the-art visuomotor policies—Diffusion Policy (DP) \cite{diffusion} and Crossway Diffusion Policy (CDP) \cite{cdp}—trained from demonstrations, and construct corresponding CI sets using the method described in Section~\ref{sec:h_x}, followed by integration of the recovery controller from Section~\ref{sec:recovery}. Experiments are conducted in three stages. In the first, baseline rollouts are performed without recovery in environments matching training conditions: \( N = \{60, 100, 120\} \) rollouts for simulation and \( N = \{20, 30, 50\} \) for real-world, each with a fixed horizon of \( T = 1300 \). The best observed success rates are used as reference. In the second stage, the same policies are re-evaluated under the same conditions but with recovery active, ensuring execution remains within \( \mathcal{S} \); performance matching or exceeding baseline confirms the presence of an execution guarantee. Finally, we test policy resilience under three OOD disturbances: OOD-I (unseen start states with mid-trajectory perturbations), OOD-II (novel distractor objects in the scene), and OOD-III (a combination of both). A policy that retains baseline-level performance when constrained to \( \mathcal{S} \), but degrades outside of it, is considered to exhibit strong safeset-based robustness.
\vspace{-0.5em}
\subsection{Experimental Setup}
\vspace*{-1em}
\begin{wraptable}[11]{r}{0.6\linewidth}
\centering
\scriptsize
\setlength{\tabcolsep}{4pt}
\renewcommand{\arraystretch}{0.95}
\caption{Task success rates under clean and cluttered conditions.}
\begin{tabular}{llcccccc}
\toprule
\textbf{Setting} & \textbf{Method} & \multicolumn{3}{c}{Clean} & \multicolumn{3}{c}{Cluttered} \\
& & N=60 & N=100 & N=120 & N=60 & N=100 & N=120 \\
\midrule
\multirow{4}{*}{Sim}
& CDP & 0.93 & 0.91 & 0.92 & 0.48 & 0.52 & 0.50 \\
& Recovery+CDP & \textbf{1.00} & \textbf{1.00} & \textbf{1.00} & \textbf{0.80} & \textbf{0.78} & \textbf{0.79} \\
& DP & 0.40 & 0.70 & 0.68 & 0.80 & 0.75 & 0.80 \\
& Recovery+DP & \textbf{0.98} & \textbf{0.99} & \textbf{0.99} & \textbf{0.93} & \textbf{0.92} & \textbf{0.92} \\
\midrule
\multirow{4}{*}{Real}
& CDP & 0.55 & 0.62 & 0.60 & 0.50 & 0.60 & 0.59 \\
& Recovery+CDP & \textbf{0.85} & \textbf{0.86} & \textbf{0.87} & \textbf{0.82} & \textbf{0.83} & \textbf{0.84} \\
& DP & 0.65 & 0.68 & 0.70 & 0.70 & 0.65 & 0.68 \\
& Recovery+DP & \textbf{0.84} & \textbf{0.85} & \textbf{0.85} & \textbf{0.86} & \textbf{0.86} & \textbf{0.86} \\
\bottomrule
\end{tabular}
\label{tab:clean_clutter}
\end{wraptable}

\paragraph{Tasks and Environments.}
For simulation, we use the \textsl{Lift} task from the RoboMimic suite. Two variants are defined: a \textbf{clean} environment identical to the original task, and a \textbf{cluttered} variant containing a visually similar distractor object (same shape, different color). We collect 110 and 120 demonstrations in the clean and cluttered settings, respectively, using a 6DOF space mouse. In the real world, we consider a \textsl{Candy Sorting} task with a Franka Emika Panda robot. The robot must pick up a colored candy and deposit it into a white bin. The clean setup contains a single object, while the cluttered setup includes three distractor candies. In both cases, the robot uses an eye-in-hand camera. We collect 40 and 45 demonstrations in the clean and cluttered real-world environments, respectively.
\begin{figure}
    \centering
\includegraphics[width=0.8\linewidth
]{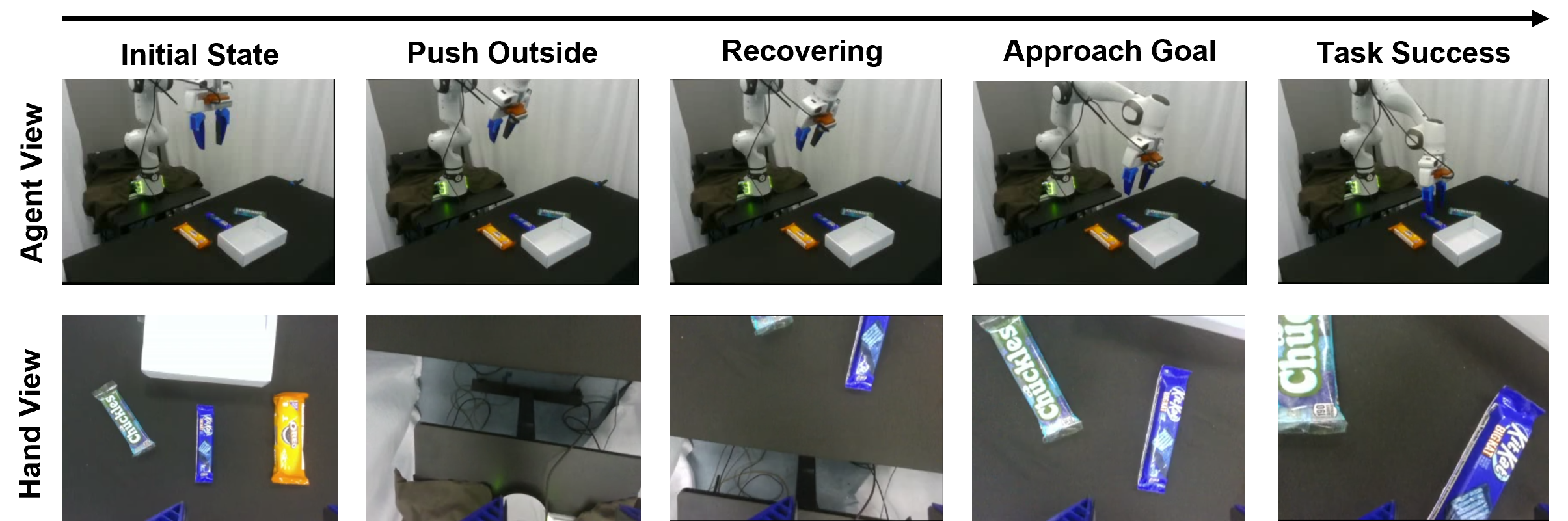}
    \caption{Policy roll outs for the real world task}
    \vspace*{-1.0em}
    \label{fig:real_recovery}
\end{figure}

\vspace*{-1.0em}

\paragraph{Policy Learning and Perception.}
We train policies using DP and CDP, each with their own perception module. The construction of \( \text{RECOG}(x) \) leverages the frozen visual encoder from each respective policy. At each candidate state, we compute the embedding of the rendered camera image and measure its Mahalanobis distance to a reference distribution built from demonstration-time embeddings. This perceptual similarity score enters into the computation of \( h(x) \) (see Eq.~\ref{eq:h_visibility}). Details are provided in Appendix~A. 

\vspace*{-1.2em}
\paragraph{Evaluation Metric.}
The primary performance metric is task success rate, measured as the fraction of episodes in which the goal state \( \mathcal{X}_{\text{goal}} \) is reached. The proposed safeset-based safety framework aims to preserve this performance level during runtime execution, even in the presence of environmental variability or perturbations.
\vspace*{-0.6em}

\subsection{Results}
\label{sec:results_analysis}
\vspace*{-1em}
The experimental results across Tables~\ref{tab:clean_clutter} and \ref{tab:ood} provide compelling and comprehensive validation of our proposed execution guarantee framework.

\vspace*{-0.5em}
\begin{wraptable}[17]{r}{0.6\linewidth}
\centering
\scriptsize
\setlength{\tabcolsep}{4pt}
\renewcommand{\arraystretch}{0.95}
\caption{Success rates under OOD conditions. Safe = with recovery controller. Baseline = clean rollout performance.}
\begin{tabular}{llcccc}
\toprule
\textbf{Env. / Method} & \textbf{Rollout} & \textbf{OOD-I} & \textbf{OOD-II} & \textbf{OOD-I+II} & \textbf{Baseline} \\
\midrule
\multirow{2}{*}{Sim / CDP} 
  & Raw & 0.83 ± 0.03 & 0.51 ± 0.02 & 0.48 ± 0.02 & 0.92 \\
  & Safe & \textbf{0.99 ± 0.00} & \textbf{0.90 ± 0.02} & \textbf{0.90 ± 0.01} & \textbf{1.00} \\
\multirow{2}{*}{Sim / CDP (Clutter)} 
  & Raw & 0.48 ± 0.01 & 0.63 ± 0.03 & 0.61 ± 0.02 & 0.52 \\
  & Safe & \textbf{0.78 ± 0.03} & \textbf{0.90 ± 0.02} & \textbf{0.86 ± 0.03} & \textbf{0.86} \\
\midrule
\multirow{2}{*}{Sim / DP} 
  & Raw & 0.65 ± 0.02 & 0.47 ± 0.02 & 0.35 ± 0.03 & 0.78 \\
  & Safe & \textbf{0.92 ± 0.03} & \textbf{0.95 ± 0.01} & \textbf{0.89 ± 0.02} & \textbf{0.99} \\
\multirow{2}{*}{Sim / DP (Clutter)} 
  & Raw & 0.59 ± 0.01 & 0.70 ± 0.01 & 0.77 ± 0.00 & 0.75 \\
  & Safe & \textbf{0.92 ± 0.02} & \textbf{0.93 ± 0.01} & \textbf{0.90 ± 0.00} & \textbf{0.95} \\
\midrule
\multirow{2}{*}{Real / CDP} 
  & Raw & 0.50 ± 0.04 & 0.52 ± 0.02 & 0.42 ± 0.05 & 0.62 \\
  & Safe & \textbf{0.82 ± 0.01} & \textbf{0.84 ± 0.01} & \textbf{0.84 ± 0.01} & \textbf{0.86} \\
\multirow{2}{*}{Real / CDP (Clutter)} 
  & Raw & 0.52 ± 0.04 & 0.58 ± 0.02 & 0.50 ± 0.01 & 0.60 \\
  & Safe & \textbf{0.79 ± 0.03} & \textbf{0.80 ± 0.02} & \textbf{0.80 ± 0.02} & \textbf{0.83} \\
\midrule
\multirow{2}{*}{Real / DP} 
  & Raw & 0.50 ± 0.02 & 0.48 ± 0.05 & 0.45 ± 0.04 & 0.68 \\
  & Safe & \textbf{0.80 ± 0.00} & \textbf{0.84 ± 0.01} & \textbf{0.83 ± 0.03} & \textbf{0.85} \\
\multirow{2}{*}{Real / DP (Clutter)} 
  & Raw & 0.50 ± 0.03 & 0.53 ± 0.00 & 0.51 ± 0.00 & 0.65 \\
  & Safe & \textbf{0.81 ± 0.01} & \textbf{0.82 ± 0.00} & \textbf{0.80 ± 0.00} & \textbf{0.86} \\
\bottomrule
\end{tabular}
\label{tab:ood}
\end{wraptable}

\textbf{Simulation Results.} In Table~\ref{tab:clean_clutter}, we observe that baseline policies achieve respectable success rates under nominal (clean) conditions (CDP: 0.93, DP: 0.70). However, performance deteriorates significantly in cluttered scenes (CDP: 0.52), revealing the brittleness of open-loop visuomotor policies when exposed to even modest perceptual variation. When constrained to their respective control-invariant (CI) safesets via our recovery controller, both policies exhibit remarkable stability. CDP consistently achieves perfect success (1.00) and DP reaches 0.99 in clean settings. More impressively, both maintain success rates above 0.86 in cluttered environments—across all rollout counts. These results demonstrate that enforcing safeset invariance is sufficient to recover or even exceed baseline performance, regardless of environmental complexity or rollout budget.

\textbf{Real-World Results.} This pattern holds in real-world deployments (Table~\ref{tab:clean_clutter}). Without recovery, policy performance is modest even under clean conditions (CDP: 0.62, DP: 0.65), and degrades further with clutter (CDP: 0.59, DP: 0.70). However, enforcing the execution guarantee via our framework significantly boosts robustness. Both Recovery+CDP and Recovery+DP surpass 0.84 in all rollout settings, without requiring retraining or access to a dynamics model. These results confirm the practicality of safeset enforcement in real physical systems, including visually ambiguous scenes and limited demonstration data.
\textbf{Out-of-Distribution Robustness.} The benefits of our framework are even more pronounced in OOD settings. In simulation (Table~\ref{tab:ood}), raw policies collapse under perturbations (e.g., DP drops to 0.35 in the OOD-I+II condition). Yet, with safeset enforcement, Recovery+DP climbs back to 0.89 and Recovery+CDP to 0.90—exceeding their own nominal baselines. This highlights the ability of our approach to absorb perceptual perturbations and restore consistent behavior.\\
A similar trend emerges in real-world OOD scenarios (Table~\ref{tab:ood}). While raw policies deteriorate sharply (e.g., CDP: 0.62 \(\rightarrow\) 0.42), recovery-enhanced variants remain robust (CDP: 0.84, DP: 0.83). These results are particularly significant, as they demonstrate generalization across real-world sensory shifts and confirm the central thesis of this paper: that execution guarantee—enforced through a control-invariant safeset and runtime recovery—translates theoretical safety into resilient, high-performance behavior across deployment settings.
\begin{figure}
    \centering
\includegraphics[scale=0.7
]{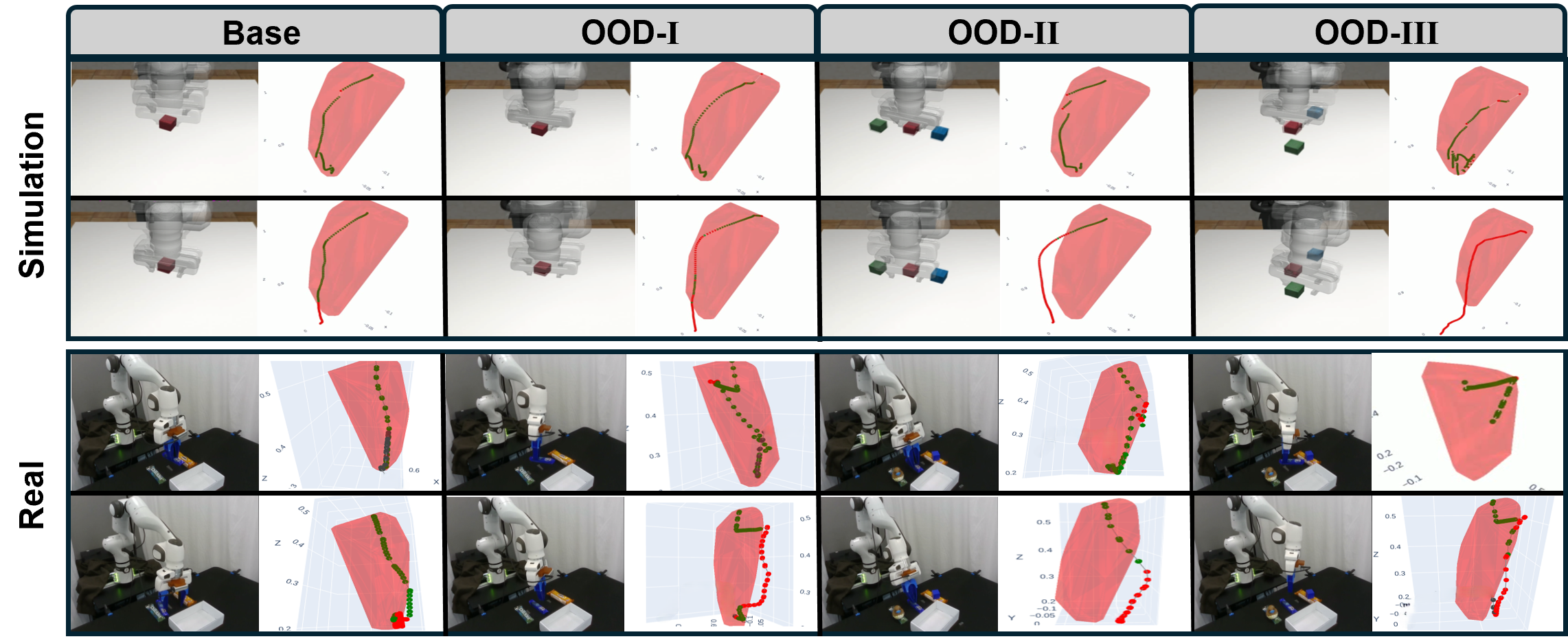}
    \caption{Rollouts in simulation and real settings under baseline and OOD conditions. Each cell shows the robot’s view (left) and safeset trajectory (right); red = outside \( \mathcal{S} \), green = inside. 
    }
    \vspace*{-1.7em}
    \label{fig:policy_rollouts}
\end{figure}
\vspace*{-1em}
\paragraph{Qualitative Observations.}
Figure~\ref{fig:policy_rollouts} provides qualitative support for these findings. Under clean conditions, both DP and Crossway DP reach the goal reliably. However, in OOD-I, II, and III, raw rollouts frequently drift outside the safeset \( \mathcal{S} \) (indicated by red markers), resulting in task failure. In contrast, trajectories using our recovery controller remain entirely within \( \mathcal{S} \) (green markers), effectively navigating clutter, novel distractors, and external perturbations. These visual outcomes reinforce our quantitative analysis: maintaining safeset invariance through our proposed execution guarantee framework leads to safe and robust behavior, even in challenging runtime scenarios.
\vspace*{-1em}
\section{Limitations and Future Work}
\label{sec:limitations}
\vspace*{-1em}
The safeset construction process is currently goal-specific and assumes demonstration data sufficiently covers perceptually consistent goal regions; generalizing this to arbitrary goal configurations or dynamic targets remains an open challenge. Additionally, the framework does not explicitly model grasp feasibility. Integrating grasp-conditioned embeddings or learning grasp-aware safe sets could improve performance, especially in cluttered scenes. The recovery controller ensures safety but can be overly conservative in constrained environments; predictive or learning-based alternatives may yield more efficient re-entry behaviors. 
\vspace*{-1.0em}
\section{Conclusion}
\label{sec:conclusion}
\vspace*{-1em}
This work presents a principled framework for certifying and enforcing safety in imitation learning through the notion of an \textit{execution guarantee}. We define execution safety in terms of control-invariant safesets constructed from demonstrations, using the policy’s own perception system to identify regions of visual competence. By leveraging a differentiable, Lipschitz-continuous safeset function and enforcing Nagumo’s condition through a recovery controller, our approach guarantees safe execution without requiring access to system dynamics or task-specific models. Extensive experiments in simulation and with a real Franka robot demonstrate the performance of the proposed work. 

\clearpage

\bibliography{references}

@inproceedings{safedagger,
  title={Sample efficient interactive end-to-end deep learning for self-driving cars with selective multi-class safe dataset aggregation},
  author={Bicer, Yunus and Alizadeh, Ali and Ure, Nazim Kemal and Erdogan, Ahmetcan and Kizilirmak, Orkun},
  booktitle={2019 IEEE/RSJ International Conference on Intelligent Robots and Systems (IROS)},
  pages={2629--2634},
  year={2019},
  organization={IEEE}
}

@inproceedings{bansal2017hamilton,
  title={Hamilton-jacobi reachability: A brief overview and recent advances},
  author={Bansal, Somil and Chen, Mo and Herbert, Sylvia and Tomlin, Claire J},
  booktitle={2017 IEEE 56th Annual Conference on Decision and Control (CDC)},
  pages={2242--2253},
  year={2017},
  organization={IEEE}
}

@article{mandlekar2021matters,
  title={What matters in learning from offline human demonstrations for robot manipulation},
  author={Mandlekar, Ajay and Xu, Danfei and Wong, Josiah and Nasiriany, Soroush and Wang, Chen and Kulkarni, Rohun and Fei-Fei, Li and Savarese, Silvio and Zhu, Yuke and Mart{\'\i}n-Mart{\'\i}n, Roberto},
  journal={arXiv preprint arXiv:2108.03298},
  year={2021}
}

@article{brunke2022safe,
  title={Safe learning in robotics: From learning-based control to safe reinforcement learning},
  author={Brunke, Lukas and Greeff, Melissa and Hall, Adam W and Yuan, Zhaocong and Zhou, Siqi and Panerati, Jacopo and Schoellig, Angela P},
  journal={Annual Review of Control, Robotics, and Autonomous Systems},
  volume={5},
  number={1},
  pages={411--444},
  year={2022},
  publisher={Annual Reviews}
}

@article{diffusion,
  title={Diffusion policy: Visuomotor policy learning via action diffusion},
  author={Chi, Cheng and Xu, Zhenjia and Feng, Siyuan and Cousineau, Eric and Du, Yilun and Burchfiel, Benjamin and Tedrake, Russ and Song, Shuran},
  journal={The International Journal of Robotics Research},
  pages={02783649241273668},
  year={2023},
  publisher={SAGE Publications Sage UK: London, England}
}

@inproceedings{cdp,
  author={Xiang Li and Varun Belagali and Jinghuan Shang and Michael S. Ryoo},
  title={Crossway Diffusion: Improving Diffusion-based Visuomotor Policy via Self-supervised Learning},
  year={2024},
  pages={16841--16849},
  booktitle={2024 IEEE International Conference on Robotics and Automation (ICRA)}
}

@inproceedings{arnab2021vivit,
  title={Vivit: A video vision transformer},
  author={Arnab, Anurag and Dehghani, Mostafa and Heigold, Georg and Sun, Chen and Lu{\v{c}}i{\'c}, Mario and Schmid, Cordelia},
  booktitle={Proceedings of the IEEE/CVF international conference on computer vision},
  pages={6836--6846},
  year={2021}
}

@inproceedings{sojib2024self,
  title={Self Supervised Detection of Incorrect Human Demonstrations: A Path Toward Safe Imitation Learning by Robots in the Wild},
  author={Sojib, Noushad and Begum, Momotaz},
  booktitle={2024 IEEE/RSJ International Conference on Intelligent Robots and Systems (IROS)},
  pages={2862--2869},
  year={2024},
  organization={IEEE}
}

@book{blanchini2008set,
  title={Set-theoretic methods in control},
  author={Blanchini, Franco and Miani, Stefano and others},
  volume={78},
  year={2008},
  publisher={Springer}
}

@article{osa2018algorithmic,
  title={An algorithmic perspective on imitation learning},
  author={Osa, Takayuki and Pajarinen, Joni and Neumann, Gerhard and Bagnell, J Andrew and Abbeel, Pieter and Peters, Jan and others},
  journal={Foundations and Trends in Robotics},
  volume={7},
  number={1-2},
  pages={1--179},
  year={2018},
  publisher={Now Publishers, Inc.}
}

@article{generalizingSafety,
  title={Generalizing Safety Beyond Collision-Avoidance via Latent-Space Reachability Analysis},
  author={Nakamura, Kensuke and Peters, Lasse and Bajcsy, Andrea},
  journal={arXiv preprint arXiv:2502.00935},
  year={2025}
}

@article{wabersich2022mpsf,
  title={Predictive Safety Filter: A Modular Safety Layer for Learning-Based Control},
  author={Wabersich, Jonas and Zeilinger, Melanie N.},
  journal={IEEE Transactions on Automatic Control},
  year={2022},
  note={Early Access}
}

@article{wabersich2023data,
  title={Data-driven safety filters: Hamilton-jacobi reachability, control barrier functions, and predictive methods for uncertain systems},
  author={Wabersich, Kim P and Taylor, Andrew J and Choi, Jason J and Sreenath, Koushil and Tomlin, Claire J and Ames, Aaron D and Zeilinger, Melanie N},
  journal={IEEE Control Systems Magazine},
  volume={43},
  number={5},
  pages={137--177},
  year={2023},
  publisher={IEEE}
}

@article{ames2017cbf,
  title={Control Barrier Function Based Quadratic Programs for Safety Critical Systems},
  author={Ames, Aaron D. and Xu, Xuan and Grizzle, Jessy W. and Tabuada, Paulo},
  journal={IEEE Transactions on Automatic Control},
  year={2017},
  volume={62},
  number={8},
  pages={3861--3876}
}

@inproceedings{fisac2018generalizing,
  title={A General Safety Framework for Learning in Uncertain Robotic Environments},
  author={Fisac, Juan P. and Akametalu, A. Koushil and Zeilinger, Melanie N. and Huang, Wenhao and Tomlin, Claire J.},
  booktitle={Proc. IEEE Conf. on Decision and Control (CDC)},
  year={2018},
  pages={2735--2742}
}

@inproceedings{alshiekh2018shielding,
  title={Safe Reinforcement Learning via Shielding},
  author={Alshiekh, Mohammad and Bloem, Roderick and Ehlers, R{\"u}diger and Kochenderfer, Mykel J. and Topcu, Ufuk and Vardi, Moshe Y.},
  booktitle={Proc. AAAI Conf. on Artificial Intelligence},
  year={2018},
  pages={2669--2678}
}

@inproceedings{berkenkamp2017safe,
  title={Safe Model-Based Reinforcement Learning with Stability Guarantees},
  author={Berkenkamp, Felix and Turchetta, Matteo P. and Schoellig, Angela P. and Krause, Andreas},
  booktitle={Advances in Neural Information Processing Systems (NeurIPS)},
  year={2017},
  pages={908--918}
}

@inproceedings{kahn2017uncertainty,
  title={Uncertainty-Aware Reinforcement Learning for Collision Avoidance},
  author={Kahn, Gregory and Villaflor, Angelika and Ding, Vitchyr Pong and Abbeel, Pieter and Levine, Sergey},
  booktitle={Proc. IEEE Int. Conf. on Robotics and Automation (ICRA)},
  year={2017},
  pages={316--323}
}

@article{xiao2023safediffuser,
  title={Safediffuser: Safe planning with diffusion probabilistic models},
  author={Xiao, Wei and Wang, Tsun-Hsuan and Gan, Chuang and Rus, Daniela},
  journal={arXiv preprint arXiv:2306.00148},
  year={2023}
}

@article{xiao2023barriernet,
  title={Barriernet: Differentiable control barrier functions for learning of safe robot control},
  author={Xiao, Wei and Wang, Tsun-Hsuan and Hasani, Ramin and Chahine, Makram and Amini, Alexander and Li, Xiao and Rus, Daniela},
  journal={IEEE Transactions on Robotics},
  volume={39},
  number={3},
  pages={2289--2307},
  year={2023},
  publisher={IEEE}
}

@article{nagumo1942,
  title={{\"U}ber die Lage der Integralkurven gew{\"o}hnlicher Differentialgleichungen},
  author={Nagumo, Mitio},
  journal={Proceedings of the Physico-Mathematical Society of Japan. 3rd Series},
  volume={24},
  pages={551--559},
  year={1942},
  publisher={J-STAGE}
}

@article{jung2024rail,
  title={RAIL: Reachability-Aided Imitation Learning for Safe Policy Execution},
  author={Jung, Wonsuhk and Anthony, Dennis and Mishra, Utkarsh A and Arachchige, Nadun Ranawaka and Bronars, Matthew and Xu, Danfei and Kousik, Shreyas},
  journal={arXiv preprint arXiv:2409.19190},
  year={2024}
}

@article{tayal2024semi,
  title={Semi-supervised safe visuomotor policy synthesis using barrier certificates},
  author={Tayal, Manan and Singh, Aditya and Jagtap, Pushpak and Kolathaya, Shishir},
  journal={arXiv preprint arXiv:2409.12616},
  year={2024}
}

@article{xiao2024abnet,
  title={ABNet: Attention BarrierNet for Safe and Scalable Robot Learning},
  author={Xiao, Wei Brandon and Wang, Tsun-Hsuan and Rus, Daniela},
  journal={arXiv preprint arXiv:2406.13025},
  year={2024}
}

@article{blanchini1999set,
  title={Set invariance in control},
  author={Blanchini, Franco},
  journal={Automatica},
  volume={35},
  number={11},
  pages={1747--1767},
  year={1999},
  publisher={Elsevier}
}

\vfill
\clearpage
\appendix

\section{Safe set Construction}
\subsection{Field-of-View Constraint}
\label{sec:appendix:fov}

The $FOV(x)$ term evaluates whether the object of interest is visible from the candidate robot state \( x \), based on a known camera projection model. Given the end-effector pose, we transform the 3D object coordinates into the camera frame and project them to 2D image-space coordinates \( (u, v) \). Visibility is assessed using both the angular deviation from the optical axis and the projection's location on the image plane. To provide a smooth and differentiable signal for learning and control, $FOV(x)$ is defined as a Gaussian-like score centered at the image midpoint \( (c_u, c_v) \), with the falloff rate determined by a spread parameter \( \sigma \). The value is clipped to zero if the projection lies outside the field of view or behind the camera (\( z \leq 0 \)). Formally,

\begin{equation}
FOV(x) =
\begin{cases}
\exp\left(-\dfrac{(u - c_u)^2 + (v - c_v)^2}{\sigma^2} \right), & \text{if } 0 < u < W,\; 0 < v < H,\; z > 0 \\
0, & \text{otherwise}
\end{cases}
\end{equation}

\begin{figure}
    \centering
\includegraphics[scale=0.6
]{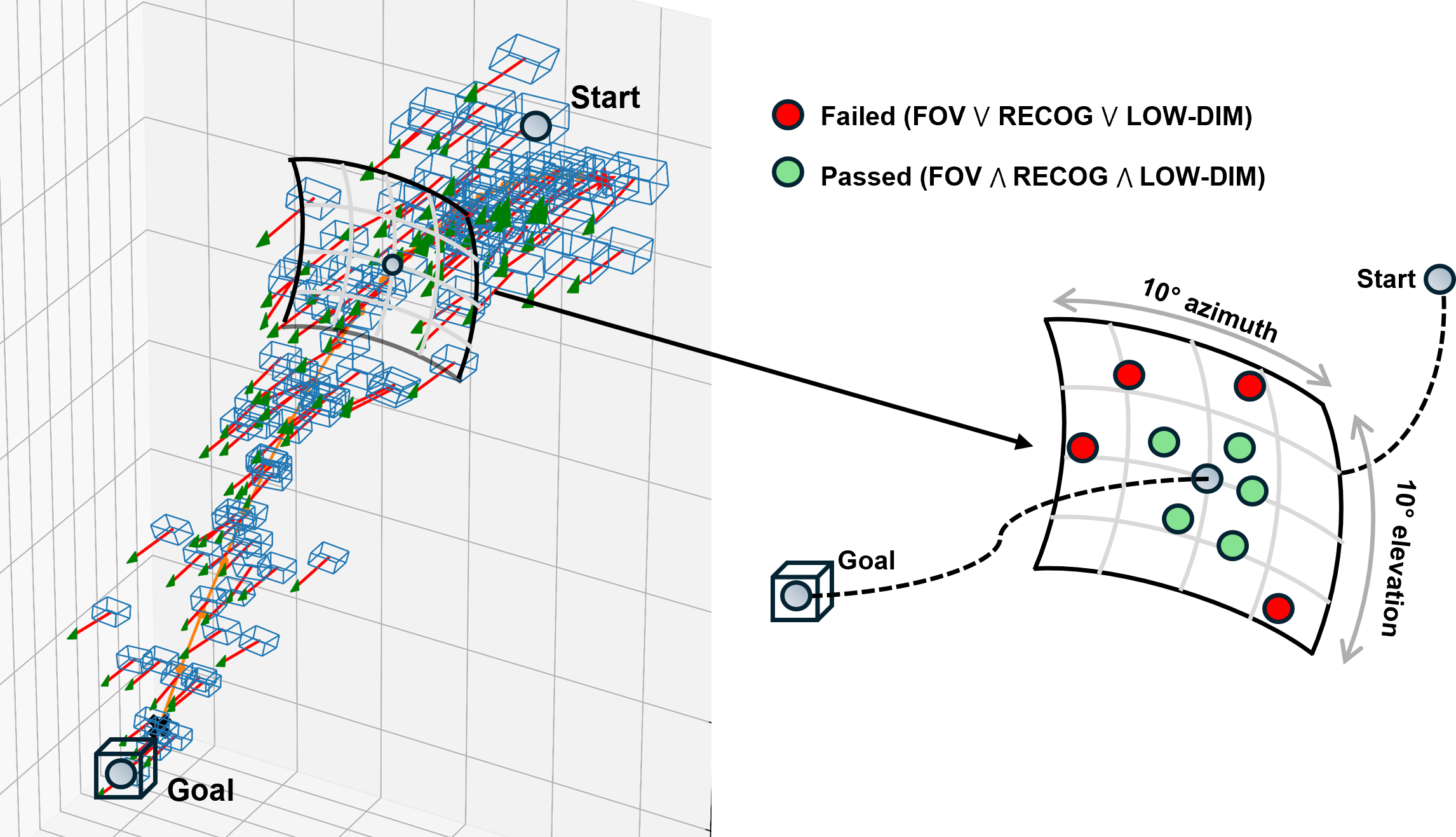}
    \caption{Candidate Pose Filtering}
    \label{fig:candidate_points_filtering}
\end{figure}

This continuous scoring allows smooth safety contours and is particularly compatible with gradient-based recovery control. A visual illustration of the image projection and scoring is provided in Figure~\ref{fig:candidate_points_filtering}.

\subsection {Recognizability Score}
\label{sec:appendix:recog}
The $RECOG(x)$ component evaluates whether the camera view rendered from a candidate robot state $x$ is perceptually similar to views observed during expert demonstrations. To enable this, we employ a novel-view synthesis pipeline based on \textbf{Gaussian Splatting}, using demonstration images and their associated camera poses to construct a 3D scene representation. During the structure-from-motion (SfM) stage of splat generation, we compute a transformation that aligns the \emph{robot coordinate frame} with the \emph{splat (scene) frame}. This alignment enables novel view rendering from arbitrary candidate robot poses $x$ by projecting them into the scene representation. The result is a synthesized image corresponding to what the robot would observe from the candidate pose.

This rendered image is passed through a frozen ResNet-18 encoder, pretrained via the corresponding imitation learning policy (e.g., Diffusion Policy or Crossway Diffusion Policy), to extract a 512-dimensional feature embedding $z(x)$. All training-time embeddings are aggregated to form a reference distribution with mean \( \mu \in \mathbb{R}^{512} \) and covariance matrix \( \Sigma \in \mathbb{R}^{512 \times 512} \). For any candidate state, we compute the Mahalanobis distance:

\begin{equation}
D^2(x) = (z(x) - \mu)^\top \Sigma^{-1} (z(x) - \mu),
\end{equation}

and define the recognizability score as a smooth exponential:

\begin{equation}
RECOG(x) = \exp\left(-\frac{1}{2} D^2(x)\right).
\end{equation}

This ensures that states visually close to the training distribution receive scores near 1, while unfamiliar observations are exponentially penalized. The output is bounded in \( (0, 1] \), providing a continuous and differentiable measure of visual familiarity. The statistics \( \mu \) and \( \Sigma \) are computed offline over the demonstration dataset, and recognizability is evaluated at runtime using the frozen policy-specific vision encoder.
\subsection{Low-Dimensional Constraint Filter}
\label{sec:appendix:ld_constraints}

To ensure geometric and task-level feasibility, a low-dimensional constraint filter is applied to restrict candidate states to physically meaningful regions. This filter operates over the robot’s low-dimensional state — typically the 7D end-effector pose — and incorporates the following three constraints:

\textbf{1. Proximity Constraint.}  
The end-effector must lie within a task-relevant distance range from the goal object. Let \( d(x_{\text{eef}}, g) = \|x_{\text{eef}} - g\|_2 \). A candidate state is retained only if:

\begin{equation}
d_{\min} \leq d(x_{\text{eef}}, g) \leq d_{\max},
\end{equation}

with bounds \( d_{\min} \) and \( d_{\max} \) estimated from empirical statistics over the demonstration trajectories, and margins added for robustness.

\textbf{2. Table Clearance Constraint.}  
To prevent collision with the workspace surface, the vertical position of the end-effector must exceed a minimum clearance:

\begin{equation}
z_{\text{eef}} \geq z_{\text{table}} + \epsilon,
\end{equation}

where \( z_{\text{table}} \) is the known table height and \( \epsilon \) is a tunable buffer.

\textbf{3. Orientation Alignment Constraint.}  
The end-effector’s approach direction is constrained to remain near the world’s downward axis. Let \( a_{\text{eef}} \in \mathbb{R}^3 \) be the local x-axis of the gripper (expressed in world coordinates), and \( a_{\text{world}} = [0, 0, -1]^\top \). We require:

\begin{equation}
\arccos\left( a_{\text{eef}}^\top a_{\text{world}} \right) \leq \theta_{\max},
\end{equation}

where \( \theta_{\max} \) is set based on the maximum deviation observed in demonstrations. A candidate is retained in the low-dimensional safeset only if all three constraints are satisfied:

\begin{equation}
\mathcal{S}_{\text{low-dim}} = \left\{ x \in \mathbb{R}^n \mid \text{Proximity, Clearance, and Orientation constraints satisfied} \right\}.
\end{equation}

All parameters \( (d_{\min}, d_{\max}, \epsilon, \theta_{\max}) \) are automatically tuned from the demonstration data, ensuring both consistency and robustness.

\subsection{Candidate State Generation for Safeset Construction}
\label{sec:appendix:candidate_generation}

To reliably approximate the underlying control-invariant set around expert demonstrations, we synthetically generate a dense collection of candidate robot states by perturbing demonstration data. This procedure ensures broader coverage of the reachable safe space beyond sparsely visited demonstration points and supports accurate safeset boundary estimation.

Given a demonstration trajectory \( \tau = \{(x_t, q_t)\}_{t=1}^T \), consisting of end-effector positions \( x_t \in \mathbb{R}^3 \) and orientations \( q_t \in \mathbb{R}^4 \) (unit quaternions), and a globally computed goal pose \( g = (g_{\text{pos}}, g_{\text{quat}}) \), we generate candidates as follows:
 
For each point \( (x_t, q_t) \), compute the displacement from the goal:
\[
d_t = x_t - g_{\text{pos}}.
\]
Convert this displacement to spherical coordinates \( (r_t, \theta_t, \phi_t) \), and sample angular perturbations \( (\Delta \theta, \Delta \phi) \) within a bounded range:
\[
\tilde{d}_t = \text{Sph2Cart}(r_t, \theta_t + \Delta \theta, \phi_t + \Delta \phi),
\]
\[
\tilde{x}_t = g_{\text{pos}} + \tilde{d}_t.
\]
 
Apply random roll-pitch-yaw offsets \( (\Delta \alpha, \Delta \beta, \Delta \gamma) \) to introduce variation in gripper orientation:
\[
\tilde{q}_t = q_t \otimes \text{Euler}(\Delta \alpha, \Delta \beta, \Delta \gamma),
\]
where \( \otimes \) denotes quaternion multiplication.
For each perturbed position \( \tilde{x}_t \), generate additional samples at the corners of a cube with fixed side length \( \delta_{\text{cube}} \):
\[
\tilde{x}_{\text{corner}} = \tilde{x}_t + \delta, \quad \text{for all } \delta \in \{-\delta_{\text{cube}}, +\delta_{\text{cube}}\}^3.
\]
Each cube sample inherits orientation \( \tilde{q}_t \).

The complete candidate set is:
\[
\mathcal{X}_{\text{cand}} = \left\{ (\tilde{x}_t, \tilde{q}_t) \right\}_{t=1}^T \cup \text{CubeSamples}(\tilde{x}_t, \tilde{q}_t).
\]

A higher density of perturbations leads to tighter approximations of the safeset, while sparser coverage may result in conservative boundaries. The final safeset is computed by filtering \( \mathcal{X}_{\text{cand}} \) through the field-of-view, recognizability, and low-dimensional constraints:

\[
\mathcal{S}_{\text{safe}} = \left\{ x \in \mathcal{X}_{\text{cand}} \mid \texttt{FOV}(x), \texttt{RECOG}(x), \texttt{low-dim}(x) \text{ satisfied} \right\}.
\]

\section{Lipschitz Continuity of the Function $h(x)$}
\label{sec:appendix:lipschitz}

We show that the safety function
\[
h(x) = \text{FOV}(x) \cdot \text{RECOG}(x),
\]
defined over a compact domain $\mathcal{X} \subset \mathbb{R}^{n_x}$, is Lipschitz continuous. This property is essential for the application of Nagumo’s theorem in constructing recovery controllers that guarantee safety.

\subsection*{Preliminaries}

Let $x \in \mathbb{R}^{n_x}$ represent a robot state (typically in SE(3) pose space), and let $h: \mathbb{R}^{n_x} \rightarrow \mathbb{R}$ be defined as a product of two perceptual modules:
\begin{itemize}
    \item $FOV(x) \in [0, 1]$ — a smooth visibility score based on camera projection geometry,
    \item $RECOG(x) \in [0, 1]$ — a smooth familiarity score derived from Mahalanobis distance in image embedding space.
\end{itemize}

We aim to prove that $h(x)$ is Lipschitz continuous over compact subsets of $\mathbb{R}^{n_x}$.

\subsection*{Definition: Lipschitz Continuity}

A function $f: \mathbb{R}^{n} \rightarrow \mathbb{R}$ is said to be Lipschitz continuous on a set $\mathcal{X} \subseteq \mathbb{R}^{n}$ if there exists a constant $L > 0$ such that for all $x_1, x_2 \in \mathcal{X}$,
\[
|f(x_1) - f(x_2)| \leq L \|x_1 - x_2\|.
\]

\subsection*{Structure of $h(x)$}

We analyze the structure of $h(x)$ as a product of two functions:
\[
h(x) = f_1(x) \cdot f_2(x),
\]
where $f_1(x) := \text{FOV}(x)$ and $f_2(x) := \text{RECOG}(x)$. It is a known result that the product of two Lipschitz functions is also Lipschitz if both functions are bounded. Let us prove the Lipschitz continuity of each component and then analyze their composition.

\subsection*{Lipschitz Continuity of $\text{FOV}(x)$}

Let $P(K, x)$ denote the projection of the 3D point into image coordinates using camera intrinsics $K$ and robot pose $x$, and let $g(u, v)$ denote the Gaussian scoring function. Then,
\[
{\text{FOV}}(x) =
\begin{cases}
g(P(K, x)), & \text{if projection is valid (i.e., in front of and within the image bounds)} \\
0, & \text{otherwise}
\end{cases}
\]

The projection function $P(K, x)$ involves smooth transformations (e.g., SE(3) motion and pinhole projection), and is Lipschitz continuous on compact subsets of pose space that avoid degenerate configurations (such as projection through the optical axis). The Gaussian scoring function $g(u, v)$ is smooth and Lipschitz on its domain.

Since both components are Lipschitz continuous (on regions where the projection is valid), the composed function ${\text{FOV}}(x)$ is \textit{piecewise Lipschitz continuous}. Thus, there exists a constant $L_1 > 0$ such that for any two valid poses $x_1, x_2$,
\[
|FOV(x_1) - FOV(x_2)| \leq L_1 \|x_1 - x_2\|.
\]

This ensures that the visibility scoring function varies smoothly with respect to the robot’s pose, which is beneficial for gradient-based optimization and control.





\subsection*{Lipschitz Continuity of $\text{RECOG}(x)$}

Let $z(x) \in \mathbb{R}^d$ denote the feature embedding of the image rendered at robot state $x$ using a frozen CNN encoder. Assume the encoder is Lipschitz, i.e.,
\[
\|z(x_1) - z(x_2)\| \leq L_z \|x_1 - x_2\|, \quad \forall x_1, x_2 \in \mathcal{X}.
\]

The recognizability score is then defined as:
\[
RECOG(x) = \exp\left( -\frac{1}{2} D^2(x) \right), \quad \text{where} \quad D^2(x) = (z(x) - \mu)^\top \Sigma^{-1} (z(x) - \mu),
\]
with $\mu$ and $\Sigma$ denoting the mean and covariance of the demonstration embeddings.

Note:
- $D^2(x)$ is a smooth quadratic form.
- The exponential function is smooth and Lipschitz on compact domains.

By composition:
\[
|RECOG(x_1) - RECOG(x_2)| \leq L_2 \|x_1 - x_2\|, \quad \text{for some } L_2 > 0.
\]

\subsection*{Product Rule: $h(x) = FOV(x) \cdot RECOG(x)$}

Let $f_1(x)$ and $f_2(x)$ be Lipschitz with constants $L_1$, $L_2$, and both bounded on $\mathcal{X}$, i.e., $|f_1(x)| \leq M_1$, $|f_2(x)| \leq M_2$ for all $x \in \mathcal{X}$.

Then for all $x_1, x_2 \in \mathcal{X}$:
\begin{align*}
|h(x_1) - h(x_2)| &= |f_1(x_1) f_2(x_1) - f_1(x_2) f_2(x_2)| \\
&= |f_1(x_1)(f_2(x_1) - f_2(x_2)) + f_2(x_2)(f_1(x_1) - f_1(x_2))| \\
&\leq |f_1(x_1)| \cdot |f_2(x_1) - f_2(x_2)| + |f_2(x_2)| \cdot |f_1(x_1) - f_1(x_2)| \\
&\leq M_1 L_2 \|x_1 - x_2\| + M_2 L_1 \|x_1 - x_2\| \\
&= (M_1 L_2 + M_2 L_1) \|x_1 - x_2\|.
\end{align*}

Hence, $h(x)$ is Lipschitz continuous with constant $L_h = M_1 L_2 + M_2 L_1$ on $\mathcal{X}$.

We have shown that:
\begin{itemize}
    \item Both $FOV(x)$ and $RECOG(x)$ are Lipschitz continuous on compact domains,
    \item Their product $h(x)$ is therefore Lipschitz,
    \item Thus, the safe set $\mathcal{S} = \{ x \mid h(x) \geq 0 \}$ has a well-defined boundary and admits gradient-based recovery conditions.
\end{itemize}

\end{document}